\documentclass[sigchi]{acmart}

\AtBeginDocument{%
  \providecommand\BibTeX{{%
    \normalfont B\kern-0.5em{\scshape i\kern-0.25em b}\kern-0.8em\TeX}}}

\setcopyright{acmcopyright}
\copyrightyear{2023} 
\acmYear{2023} 
\setcopyright{rightsretained} 
\acmConference[CHI EA '23]{Extended Abstracts of the 2023 CHI Conference on Human Factors in Computing Systems}{April 23--28, 2023}{Hamburg, Germany}
\acmBooktitle{Extended Abstracts of the 2023 CHI Conference on Human Factors in Computing Systems (CHI EA '23), April 23--28, 2023, Hamburg, Germany}\acmDOI{10.1145/3544549.3585715}
\acmISBN{978-1-4503-9422-2/23/04}

\begin{document}

\title{Exploring the Use of Collaborative Robots in Cinematography}

  \author{Pragathi Praveena}
  \authornote{Both authors contributed equally to this research.}
\affiliation{%
  \institution{Department of Computer Sciences,\\ University of Wisconsin--Madison,}
  \city{Madison}
  \state{WI}
  \country{USA}
  }
\email{pragathi@cs.wisc.edu}

  \author{Bengisu Cagiltay}
  \authornotemark[1]
\affiliation{%
  \institution{Department of Computer Sciences,\\ University of Wisconsin--Madison,}
  \city{Madison}
  \state{WI}
  \country{USA}
  }
\email{bengisu@cs.wisc.edu}

\author{Michael Gleicher}
\affiliation{%
  \institution{Department of Computer Sciences,\\ University of Wisconsin--Madison,}
  \city{Madison}
  \state{WI}
  \country{USA}
  }
\email{gleicher@cs.wisc.edu}

\author{Bilge Mutlu}
\affiliation{%
  \institution{Department of Computer Sciences,\\ University of Wisconsin--Madison,}
  \city{Madison}
  \state{WI}
  \country{USA}
  }
\email{bilge@cs.wisc.edu}

\renewcommand{\shortauthors}{Pragathi Praveena, Bengisu Cagiltay, Michael Gleicher, Bilge Mutlu}


\begin{abstract}
Robotic technology can support the creation of new tools that improve the creative process of cinematography. It is crucial to consider the specific requirements and perspectives of industry professionals when designing and developing these tools. In this paper, we present the results from exploratory interviews with three cinematography practitioners, which included a demonstration of a prototype robotic system. We identified many factors that can impact the design, adoption, and use of robotic support for cinematography, including: (1) the ability to meet requirements for cost, quality, mobility, creativity, and reliability; (2) the compatibility and integration of tools with existing workflows, equipment, and software; and (3) the potential for new creative opportunities that robotic technology can open up. Our findings provide a starting point for future co-design projects that aim to support the work of cinematographers with collaborative robots.

\end{abstract}




\begin{teaserfigure}
\centering    \includegraphics[width=\textwidth]{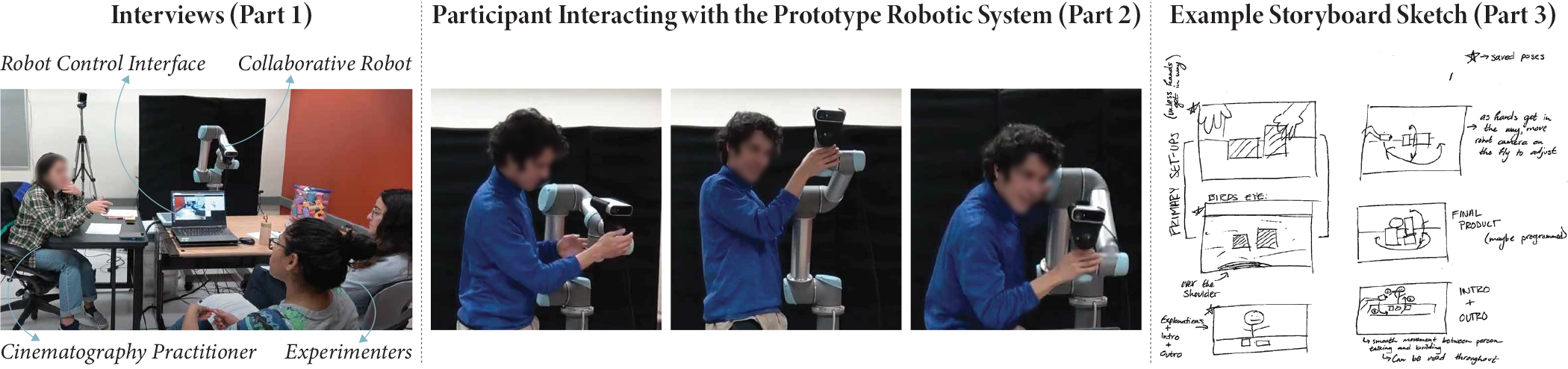}
  \caption{We interviewed three cinematography practitioners to investigate the design space of robot-supported cinematography. The study was divided into three parts. (Left) \textit{Part 1:} Illustration of the interview setup. (Middle) \textit{Part 2:} A participant interacting with our prototype robotic system while describing how they envision such technology in media production. (Right) \textit{Part 3:} An example storyboard sketched by a participant that incorporated the robot into their filming process.}
  \Description{There are five images in this figure. The left panel has one image that illustrates a room with a collaborative robot arm and robot control interface. In this room, there are three people sitting: one cinematography practitioner is describing their design preferences to two experimenters. In the center panel, there are three images that illustrate a sequence of actions that a cinematography practitioner is taking while interacting with a collaborative robot via direct physical contact during our user study. The right panel shows a sketch that was drawn by a cinematography practitioner. The sketch includes six tiles that illustrate a storyboard for robot-supported cinematography. One tile illustrates a primary setup, and a tile below that illustrates the bird's eye view of the primary setup. The tile below includes shots for explanations, intro, and outro. Three tiles on the right illustrate views for showing a product and angles the robot might take while shooting them.}
  \label{fig:teaser}
\end{teaserfigure}

\maketitle

\section{Introduction}
In recent years, the use of robotic and automation technology has grown in various industries, including the film industry, as a way to improve efficiency, lower costs, and increase quality in tasks such as camera operation, lighting, and set design.
Different types of robotic systems have been employed for these purposes, such as robotic camera systems that utilize pre-programmed motion trajectories to achieve special filming effects~\cite{mrmc} and camera robots that specialize in high-speed and precision moves~\cite{arcade}.
Drones are also becoming increasingly popular for media production due to their maneuverability and the unique perspectives they offer. The potential uses of robotic technology in media production depend greatly on the specific form factor and capabilities of the robot. In this study, we focus on the potential of collaborative robot arms (also known as \textit{cobots}) for cinematography.

Cobots are designed to work safely alongside human workers in shared, collaborative workspaces. 
Cobots have been rarely used for production of media, but they offer reasonable precision and repeatability while also being safe to interact with, which opens up new design opportunities. To gain a better understanding of how such emerging technology could be used in media production, we conducted exploratory interviews with three cinematography practitioners. We interviewed them about their prior experience in cinematography, demonstrated a prototype robotic system that showcased some unique capabilities enabled by a cobot, and asked them to create a storyboard incorporating the cobot in their filming process. In this paper, we report the thematic findings from our interviews, which provide insights into how practitioners envision the use of cobot technology in media production. Our main research question is: \textit{``How do cinematography practitioners perceive the use and application of collaborative robots in media production?''}

\section{Related Work}

The use of robots for cinematography in the media production industry has been increasing in recent years. Industrial robots have traditionally been used for high-speed and precise camera control and to create special effects~\cite{mrmc, arcade, sisu}. Drones are used for aerial shots~\cite{dji, autel}, enabling creative shots that would otherwise be impossible due to cost, difficulty, or safety concerns for cinematographers. 
%
Additionally, research in robotics has explored various ways to support cinematographers. \citet{hajjaj2021adoption} found that camera operators' attitudes towards robots in the film industry improved significantly when robotic cameras and dollies are perceived as social robots by their users. 
\citet{yamamoto2022photographic} used collaborative robots to assist photographers in setting up their lighting configurations.
To facilitate the ability to take cinematic shots, \citet{pueyo2020cinemairsim} simulated a cinema camera within a drone simulator, allowing the simulator's parameters to be adjusted with the quality of the camera. \citet{christie2016automated} and \citet{gschwindt2019can} developed techniques to support automated cinematography with drones. These autonomous drones leveraged cinematographic principles to automatically select viewpoints during moving shots (\textit{e.g.}, following a moving actor) and made intelligent decisions for viewpoints to support the cinematographer in real-time. \citet{hosl2019understanding} and \citet{galvane2018directing} have explored the design of user-friendly camera control interfaces to support the cinematographer's experience, including designing for gimbals, robots, and drone interfaces. These designed interfaces can support creativity, quality, and a sense of control for cinematographers. 
Finally, robot applications for filming are not reserved for professional settings but can also be designed for amateurs for entertainment and to support creativity. \citet{sugimoto2009gentoro} used handheld projectors augmented with a robotic character to support children's creativity and storytelling skills and \citet{zeglin2014herb} used a robot in a live theatrical performance. Overall, previous research suggests that robots can support the creative processes of cinematographers in various roles and use cases.

Existing robotic camera systems tend to use industrial robots that are more precise, faster, and able to handle heavier payloads compared to collaborative robots. However, for some reduction in precision, speed, and payload capacity, cobots offer benefits such as safety, a smaller footprint, user-friendly interfaces, and lower costs. These unique capabilities present an exciting but under-explored design space for developing tools that support cinematography.

\section{User Study}


\subsection{Procedure}
We conducted two-hour interviews with three cinematography practitioners. Our study procedure started by briefing the participant about the study goals, duration, and structure. After obtaining informed consent, facilitators started the audio and video recording. The study had three parts: (1) the participant was interviewed about their experience in cinematography, past projects, and roles; (2) facilitators demonstrated the features of a prototype robotic system called \textit{Periscope} and allowed the participant to free-play with the system; and (3) the participant sketched a scenario and process for robot-supported filming. Below, we detail each part of the protocol, and include our prompts and interview questions. 

\subsubsection{Part 1: }
Our first step was to interview participants to gain an understanding of their expertise, roles, support, technology, and past filming projects. Later, we asked the participant to \textit{``pick the project you are most comfortable and knowledgeable about and briefly talk about your experience and skills.''} 

\subsubsection{Part 2: }
We demonstrated the robot's capabilities to the participants and allowed them to free-play while discussing use cases and limitations of the robot (study setup in Figure \ref{fig:teaser}).
We turned on the \textit{Periscope} web interface and introduced the robot: \textit{``This is a collaborative robotic arm. It is safe to be around and it can also be quite precise in how it moves. We are going to control this robot using the interface that you see on the laptop.''} Later, we introduced three panels on the interface: 
(1) \textbf{Live feed} from the camera that is on the end of the robot arm. 
(2) \textbf{3D simulated view} of the robot in its current state and the table in front. 
(3) \textbf{Control panel} that can be used to control the robot. 

Next, we provided a brief overview of the robotic system and asked participants to share their opinions on how they envision such technology in media production. We demonstrated four system capabilities and asked the participant to free-play: 
(1) \textbf{Physically posing} the robot to get a specific view either using a handle on the camera to move it or directly moving the joints of the robot.
(2) \textbf{Mouse control} for setting views by \textit{clicking} on the camera feed, orbiting around an object in focus by \textit{clicking and dragging}, and moving the camera closer or further by \textit{scrolling}.
(3) \textbf{Automatic hand tracking} by the robot that makes autonomous adjustments to maintain the user's right hand in the camera view. 
(4) \textbf{Reset} where the robot moves to a previously saved position.

Additionally, we described system capabilities that were not demonstrated in the study: 
(1) \textbf{Video conferencing:} The interface can be accessed from a different physical space to control the robot. All features except physically posing the robot can be used remotely. 
(2) \textbf{Support for multiple people:} Multiple people can access the robot at once with their own interfaces.  
(3) \textbf{Collision detection:} The robot is aware of its surroundings, and avoids hitting known objects like the table or people around the robot.
Finally, we asked future-facing questions, such as how the robot's capabilities apply to their experience, how the robot's capabilities might be useful or not in their domain, and if they had any suggestions to improve the system to make it more useful in their domain.

\subsubsection{Part 3: }
We provided participants a scenario of filming ``how-to'' videos of someone assembling a complex artifact. The video would be 10--20 minutes long and meant for publishing on YouTube. We added that the production style was small and scrappy with some post-production allowed, and the production team was mostly distributed with a low budget and a few days to complete the shooting. Finally, we described that the hypothetical assembly demonstration would happen in an indoor, tabletop space (similar to the study setup) and there was access to the robot and its interface. 
%
We asked participants to create a storyboard for the short video and roughly sketch the details based on this scenario (\textit{e.g.}, Figure \ref{fig:teaser}).
Participants were asked to specify the shots or views they wanted the robot to take, what would the robot be doing, and how it ties to cinematic principles. We followed up with questions that focused on capabilities the robot needs to accomplish such shots and how they would want to control the robot to achieve these shots. 

\subsection{Participants}
Participant recruitment was conducted using mailing lists targeting senior or graduate media production students at the University of Wisconsin--Madison. Participants were selected based on their experience and knowledge in the media industry, as summarized below. They received \$25 compensation for their participation. The first two authors facilitated all sessions. 

\textbf{Participant 1 }(P1, age 22, Male) is a co-founder of a feature film production company and has six years of experience working in television. P1 has prior experience in the roles of director, cinematographer, and producer on several independent films, as well as floor management at a broadcasting agency. P1 has produced three short films over the last four years, worked as the cinematographer for four short films, and served as director for two short films. P1's early career included shooting music videos, weddings, and short documentaries for local organizations such as private businesses, schools, hospitals, and dog shelters. 
%
%

\textbf{Participant 2 }(P2, age 22, Female) has three years of experience in media production. P2 currently works at a production company where they make mini documentaries and commercials and is a videographer for the university Division of the Arts. P2 also does freelance work documenting events around the local community. P2 has worked as first camera operator on multiple dance films and was first camera assistant on a short film that took multiple years to create and is now being entered into festivals.
%

\textbf{Participant 3} (P3, age 26, Male) self-described themselves as proficient in nearly every stage of media production, including acting, casting, lighting, sound recording, filming, and editing. 
P3 has worked as a commercial videographer with seven years of experience making ads for clients such as fitness centers, universities, and the hospitality industry. 

\subsection{Analysis}
The first two authors conducted qualitative analysis on the data collected from the interviews to identify themes and patterns in participant responses, following a reflexive thematic analysis approach~\cite{ braun2019reflecting, terry2020reflexive, adams2008qualititative}. The first two authors were familiarized with the data through conducting the interviews, conducted the analysis using Nvivo 12~\cite{nvivo} to generate an initial codebook, discussed the themes, and organized the findings that are reported in this paper. 

\section{Results}
We report our findings in three categories: (1) Ability to meet cinematographers' needs; (2) Compatibility with existing processes; and (3) Potential to open up new capabilities.


\subsection{Ability to meet cinematographers' needs}

\subsubsection{Requirements for cost, quality, and mobility can constrain technology choices}
The cost of equipment was acknowledged by all participants as a critical aspect. P1 explained that equipment is \textit{``almost always rented''} for single-camera narrative productions, while a TV news show that they worked on, which regularly produced content, owned the equipment. P1 also pointed out the relationship between budget and time, saying that \textit{``the more money you have, the more time you have''} for production. P2 also highlighted the connection between budget, crew, and technology, stating that \textit{``The budget usually impacts who can be hired and who's hired depends on what technology they have. So if you're paying for a more expensive cinematographer, they're gonna have more equipment and more expensive equipment.''} P3 envisioned renting a robotic system, saying, \textit{``I can definitely see a studio...that does commercial shoots [with the robot], buying one and just renting it out.''} P1 mentioned the possibility of individuals having their preferred camera equipment or other attachments and wanting to attach it to rented equipment using standard adapters.


The audio quality of recordings was important to the participants' work, and all participants acknowledged the white noise from the robot's fans as a concern. P1 stated that \textit{``noise is an important thing''} and that with the robot system, they would need to \textit{``re-record that audio separately.''} P1 also mentioned that this is not ``ideal'' as it would \textit{``cost a lot of time.''}
P3 also noted that the vibration generated by the robot could affect the quality of the video, saying, \textit{``There’s a lot of vibration, I can kind of tell looking at the video. At least I can feel it, that’s for sure.''}

Participants also discussed the need for mobility. The mobility of the robot was appreciated by the participants; for example, P3 stated, \textit{``It’s on wheels. It’s just sort of like a dolly with extra steps.''} P2 described how, in the context of filming dancers, \textit{``during each shot, the only movement is like tilting the camera up or down or panning it left and right to accommodate the dancers movements, but in between shots it could range throughout the whole space.''} P1 compared the robot to hand-held gimbals, which are convenient when going to the location for shooting because they are compact enough to \textit{``put it in a case and take it off to the middle of the desert.''} P1 also mentioned that this feature of hand-held gimbals makes it more affordable to shoot on location compared to \textit{``rent[ing] out studios and build[ing] sets.''}

\subsubsection{Robots can support creativity in a cinematographer's processes} The participants expressed that the robot would be able to support capturing visually interesting shots. P1 described that such shots would \textit{``look super cool and beautiful.''} P2 described how they normally use tripods in their work, stating that they \textit{``move the tripod around the space and move it up down...to get a shot that looks interesting,''}, and explained through sketches how this can be applied with the support of the robot. 
P1 imagined how the hand-tracking feature could be utilized for creative shots: \textit{``So if it [the robot] knew where my hands were, and it knew that I was doing something particularly intricate with my hands, and it could respond to that and get a smooth...shot that looks intentional.''} P3 emphasized the appeal of using the robot to get shots that \textit{``would be different from what anybody else is doing right now,''} and added that \textit{``it [the robot] would be worth it for people just to make it look so interesting. Like who is doing this camera work?''}



\subsubsection{Lack of trust towards the reliability of the robot can impact its use} Participants had concerns about the potential lack of trust from industry professionals towards the reliability of the robot. P2 and P3 expressed concerns about the robot's reliability in high-stakes scenarios where there is only opportunity for one or two takes. P3 said that it is \textit{``difficult to pitch to somebody--trust me, it will track you--''} especially in shots where \textit{``you only have two or three takes. It kind of kills the mood to just say `stop'. Sorry, we misprogrammed it. Give us a few minutes.''} Similarly, P2 said that \textit{``most people in the industry wouldn't trust the robot and they would trust someone with like 10 years of experience to be able to try to get that one take or two takes.''} P1 was worried about conflicts in decision-making, saying \textit{``I don't want the robot making its own decisions for me''} except for \textit{``how to keep the camera steady.''} P1 added that they would \textit{``want to be in charge of''} how the shot looks like.
P1 and P3 suggested contingency plans where they can take over in trust-sensitive situations, such as P3's idea of \textit{``a manual override with a hand.''} P1 said that they \textit{``can forgive a robot for not being able to figure it out...when I'm trying to explain,''} as it also happens with people. However, with people they \textit{``can always just take the camera and show them.''}

\subsection{Compatibility with existing processes}


\subsubsection{Automation can be an advantage for cinematographers who plan ahead} All participants noted that using the robot to automate and repeat shots with precision and consistency was beneficial. 
P1 and P2 described how repeatability is important for continuity with multiple takes, with P1 stating that the robot would be helpful to \textit{``get the exact same shot every time so that you can basically build multiple layers in post[production].''}
%
However, the participants also acknowledged the need for planning and
preparation when using automation. Planning can vary depending on the cinematographer's habits and the production context. P2 discussed a dance film where planning is limited since \textit{``it's not a choreographed dance. It's all gonna be kind of improvised...we don't necessarily have a shot list''}, and raised concerns about how automation could be handled in this context: \textit{``Would you be able to make a new trajectory on the spot or would that have to be pre-programmed like days in advance?''}
In contrast, P3 mentioned how their financial constraints for a short film impacted their planning habits and how automation could be useful in their context:
\textit{I usually just plan everything out so meticulously that when I get to shoot the shot, I really don't experiment at all...I just shoot exactly what I storyboarded and it makes it easier for editing. I just do that because of financial constraints.''}

\subsubsection{Controlling and communicating with the robot can be multimodal} All participants emphasized the importance of having tactile input to control the robot, with P1 stating that \textit{``the more tactile you get the input, the better.''} Similarly, P3 discussed the use of a joystick or other manual input to program camera movements and suggested that the ability to program movements at the frame level would be beneficial. P3 also highlighted the importance of being able to refine the programmed movements by smoothing out any \textit{``wobbles''} and control the acceleration and deceleration profile of the robot because \textit{``if you don't do that, then it's just very jarring.''} 

P1 discussed using speech as a way to communicate instructions to a camera operator, such as \textit{``I want a tracking medium shot, go back 20 feet, take a left,''} which could also be an effective way of communicating with a robot. P3 suggested using the hand-tracking feature to communicate with the robot through gestures and movements: \textit{``It would sort of work like a theremin...where you just...played in the air.''} P2 discussed the use of hand signals and earpieces as a means of communication between the camera operator and the director. However, P2 expressed concern that speech or gestures may not be precise enough to communicate with the robot, and feared that this lack of precision may lead to the robot not knowing \textit{``how much to go up or down,''} leading to the camera overcompensating.

\subsubsection{Familiarity breeds acceptance for the robot's orbiting capability} All participants expressed their support for the potential uses of a robot with the ability to orbit and compared it to other existing technologies like a gimbal. P1 noted, \textit{``Even with my gimbal, I'm never really going to get the best orbit that you guys have here.''} P2 had a similar perspective: \textit{``Let's say I have a gimbal...there's only so much my hand can...reach''} and described that the robot has \textit{``the ability to kind of go around.''} P3 described that \textit{``the ability to track [stay pointed at] that thing that it's orbiting...would be incredibly useful...probably more than anything else.''} 

\subsection{Potential to open up new capabilities}

\subsubsection{Cobots can transcend traditional automation and collaborate with humans} The participants envisioned how cobots could provide opportunities for humans and robots to work together towards co-production of media. P1 suggested the ability to make simple real-time adjustments to the robot's path through direct physical manipulation or other inputs. P1 wanted the ability to \textit{``just take it [the robot] and move it or...use some sort of other tactile input...let go and it [the robot] goes right back to whatever it was doing.''} P2 envisioned a more complex scenario where the robot captured audio using a boom pole while maintaining the pole pointed at the subject and avoiding the microphone from being in the shot. P2 said that they personally do not like this job and \textit{``would love to have a robot on set that could...hold the boom mic for me and move it around and get the best audio.''} 

\subsubsection{Cobots can advance remote working capabilities for media production} 
Participants envisioned the robot to be a useful tool for supporting remote work in media production. P1 described that \textit{``almost every phase of media production can be done from anywhere''} except for the actual filming. Remote capabilities for filming are limited and typically requires being physically present. For example, P3 described the process of filming a documentary on a person rowing around the world, but did not envision using tools for remote cinematography, instead opting to drive or fly and meet the subject in-person to transfer footage and conduct interviews. 
%
%

P1 suggested that remote working capabilities may be beneficial for a role such as visual effects supervisor who needs to travel to the filming location, and make sure that what’s captured in camera on-set can be used in post production\footnote{An important task that a VFX supervisor does is use probes to capture lighting conditions on set in order to recreate it in the CG environment.}. P1 also suggested remote working capabilities to reduce the number of crew on set for scenarios that might warrant that, for instance, intimate scenes. P3 envisioned a scenario where cinematographers based in Los Angeles could remotely operate a camera located near a \textit{``great engine engineer in Cincinnati,''} in order to conduct an interview. This would be facilitated by a local rental company, who would provide and set up the robotic camera. P3 thought that this approach would be less expensive than traditional methods, and would allow for rapid remote interviews across the country.

P1 and P3 raised concerns about the lack of spatial awareness when controlling a robotic camera remotely from a different location, especially if their only view was through the camera. P1 emphasized the need to understand the remote space where the robot was filming because it was important for \textit{``being creative and figuring out what's possible.''} Similarly, P3 stated that \textit{``even...if you're looking through a great camera, you just don't notice something until you see with your own eye.''} P1 suggested using a \textit{``360 camera on a pole behind it [robot]''} to provide a panoramic view of the space. P1 additionally suggested that remote control of the camera should be integrated with a tactile input device such as a virtual camera rig\footnote{Example virtual camera rig: \url{https://80.lv/articles/using-virtual-cameras-in-gamedev-filmmaking/}} that consists of a shoulder rig with an attached VR controller.

%

%


%
 

\subsubsection{Cobots can provide technological support for human-human collaboration} P1 and P2 recognized the potential for improving human-human collaboration using robotic cameras. P2 discussed the benefits of being able to record and replay movement on the robot for improved rehearsal and communication between the camera operator and the director, which could lead to faster and more precise execution of shots. P2 proposed a rehearsal scenario where the camera operator and the director collaborate in real-time, adjusting the camera movement until the desired shot is achieved: \textit{``You [the cinematographer] could collaborate with the director and be like--Oh, is this the shot you're envisioning?--and then the director might be like--No, try moving it this way.''} The movement can then be saved and repeated on the day of the shoot.
P1 and P2 also envisioned remote human collaboration scenarios where multiple people could remotely access and operate a robotic camera at the same time, to storyboard and prototype shots. P1 suggested that this could be done in conjunction with virtual reality headsets: \textit{``Somebody in the UK and somebody in Portugal, somebody here, all put on the headset, and then there's some real cameras somewhere...And then you're all working collaboratively and...virtually grabbing the knob [attached to the robot] and moving it around.''}
\label{sec:human}

\section{Discussion}

We found that from the perspective of cinematography practitioners, the \textit{adoption} and \textit{utilization} of technology could be impacted by various factors, such as: (1) the ability to meet cost, quality, mobility, creativity, and reliability requirements; (2) the compatibility and integration of tools with existing workflows, equipment, and software; and (3) the potential for new creative opportunities that the technology can open up. While the first two factors are essential practical considerations for researchers to keep in mind, research often centers around the third aspect---\textit{the potential for new creative opportunities}. In light of this, we propose four key takeaways.


\subsection{Takeaways}

\subsubsection{Re-imagining robot precision and repeatability} The precision and repeatability of robots have previously been utilized to maintain continuity during editing and capture multiple takes for visual effects. However, these features can be applied in new ways, such as \textit{recording and documenting camera motion} created by users during rehearsal and \mbox{co-creation}. In a rehearsal scenario, the robot enables adjustments to recorded camera movements until the desired shot is achieved. In a co-creation scenario, multiple people can control a robotic camera for storyboarding and prototyping shots. Additionally, robots can ease the transition from planning and collaboration to execution by saving and replaying captured movements on the day of the shoot. Another promising avenue for further exploration is the encoding of expert knowledge about camera motion into robot motion. This technology could be utilized, for instance, to provide real-time guidance and feedback to non-experts, thus facilitating the learning of camera techniques. 

\subsubsection{Leveraging remote camera actuation} Robotic teleoperation for manipulation tasks has been extensively studied, but less so for camera actuation in remote spaces. The increasing use of drones highlights the potential of remote camera control. On the other hand, cobots offer unique opportunities for filming, given their different mobility and degrees of freedom compared to drones. Unlike industrial robots that may require safety barriers, cobots can be used to safely move cameras remotely, making them suitable for a variety of applications. This technology has the potential to democratize access to remote experts, provide creators with greater flexibility to work within financial, time, and space constraints, and reduce crew requirements, especially in scenarios such as COVID-19 or intimate scenes. To ensure the usability of the system, it is necessary to address potential issues regarding the remote user's lack of spatial awareness.

\subsubsection{Co-producing media with a robot} 
Robots can act in a supportive role by complementing human abilities. For example, a robot can point at an object while a human moves the camera along a trajectory, or a robotic camera can avoid capturing unwanted objects in its view, such as tripods or crew members. In addition, robots can take on secondary tasks such as holding a boom microphone for the cinematographer. Such co-production of media could result in visually compelling and distinctive shots. The capacity for real-time responsiveness is essential for successful co-production with robots. This responsiveness requires users to be able to quickly edit robot movements, program the robot in-situ, and and use various communication methods to interact with the robot in real-time. While there are potential benefits to a system that can respond to dynamic changes, concerns remain about establishing such a system's reliability, as well as the possibility of the system encroaching upon the creative choices of its user.

\subsubsection{Designing robotic tools for cinematography} 
Various production types impose certain demands and constraints on the final product, which in turn shapes the technology required and developed to support the production process. We identified four main categories of design decisions when developing tools. These include (1) the broader ecosystem that the robot-assisted processes operate within; (2) the capabilities of the hardware platform (such as degrees of freedom, navigation abilities, and sensors); (3) the autonomous behaviors that improve camera control; and (4) the design of the interface for communicating with the robot which may vary depending on the user's level of control required. Designing systems that utilize cobots to support creative processes is a unique application that differs from the more traditional use of robots in manufacturing environments. These robotic systems for creative work should allow experimentation, exploration, and the discovery of new opportunities by users. Their design process should draw from prior research on tools that aid creative work, such as \citet{Shneiderman2007creativity}. For instance, one design principle that \citet{resnick2005design} recommends is: \textit{Design with low thresholds, high ceilings, and wide walls}. This implies that systems for creative work should be easy for beginners to start using, yet provide advanced functionality for experts, and have a broad range of functionality to support exploration.

\subsection{Limitations and Future Work} Our work serves as an initial step towards understanding the design of tools that utilize cobots to support cinematographers' practices. We conducted a short qualitative study to understand the perspectives of practitioners. However, our study had certain limitations. The demonstrations of prototype robot capabilities may have limited participant discussions to the observed capabilities. However, we found that participants used the demonstrated capabilities as a starting point for brainstorming and exploring new ideas, ultimately leading to interesting discussion about new and alternative cinematography processes. We also note that our work is limited by the participants included in our interviews, \textit{i.e.}, the limited sample size and diversity in professional experience. In future work, we plan to design interactive prototypes based on the insights we learned from these exploratory interviews and we plan to follow up with the practitioners that participated in our study for co-design sessions. We also plan to recruit more practitioners with diverse cinematography experiences to support our design process. 


\begin{acks}
We thank Yeping Wang for his assistance in setting up the study. This work was supported by a NASA University Leadership Initiative (ULI) grant awarded to the UW-Madison and The Boeing Company (Cooperative Agreement \#80NSSC19M0124) and an NSF award 1830242.
\end{acks}

\bibliographystyle{ACM-Reference-Format}
\bibliography{sample-base}

\end{document}